\documentclass[journal]{IEEEtran}
\usepackage{cite}

\ifCLASSINFOpdf
  \usepackage[pdftex]{graphicx}
\else
\fi
\usepackage{amsmath}
\usepackage{algorithmic}

\usepackage{array}
\usepackage{url}

\newcommand{\degree}{\ensuremath{^\circ}}
\usepackage[table,xcdraw]{xcolor}
\usepackage{multirow}

\begin{document}
%
\title{Autonomous Driving in the Lung\\using Deep Learning for Localization}
%
%
%

\author{Jake~Sganga,~\IEEEmembership{Student Member,~IEEE,}
        David~Eng,~\IEEEmembership{Student Member,~IEEE,}
        Chauncey~Graetzel,~\IEEEmembership{Member,~IEEE,}
        and~David~B.~Camarillo,~\IEEEmembership{Member,~IEEE}
\thanks{J. Sganga and D. B. Camarillo are with the Department
of Bioengineering, Stanford University, Stanford,
CA, 94305 USA e-mail: (dcamarillo@stanford.edu)}
\thanks{D. Eng is with the Department of Computer Science, Stanford University.}
\thanks{C. Graetzel is with Auris Health, Inc. (A Johnson and Johnson subsidiary)}}
\maketitle

\begin{abstract}
Lung cancer is the leading cause of cancer-related death worldwide, and early diagnosis is critical to improving patient outcomes. To diagnose cancer, a highly trained pulmonologist must navigate a flexible bronchoscope deep into the branched structure of the lung for biopsy. The biopsy fails to sample the target tissue in 26-33\% of cases largely because of poor registration with the preoperative CT map. To improve intraoperative registration, we develop two deep learning approaches to localize the bronchoscope in the preoperative CT map based on the bronchoscopic video in real-time, called AirwayNet and BifurcationNet. The networks are trained entirely on simulated images derived from the patient-specific CT. When evaluated on recorded bronchoscopy videos in a phantom lung, AirwayNet outperforms other deep learning localization algorithms with an area under the precision-recall curve of 0.97. Using AirwayNet, we demonstrate autonomous driving in the phantom lung based on video feedback alone. The robot reaches four targets in the left and right lungs in 95\% of the trials. On recorded videos in eight human cadaver lungs, AirwayNet achieves areas under the precision-recall curve ranging from 0.82 to 0.997.
\end{abstract}


%
\IEEEpeerreviewmaketitle

\section{Introduction}
\begin{figure}[thpb]
  \centering
  \includegraphics[width=0.45\textwidth]{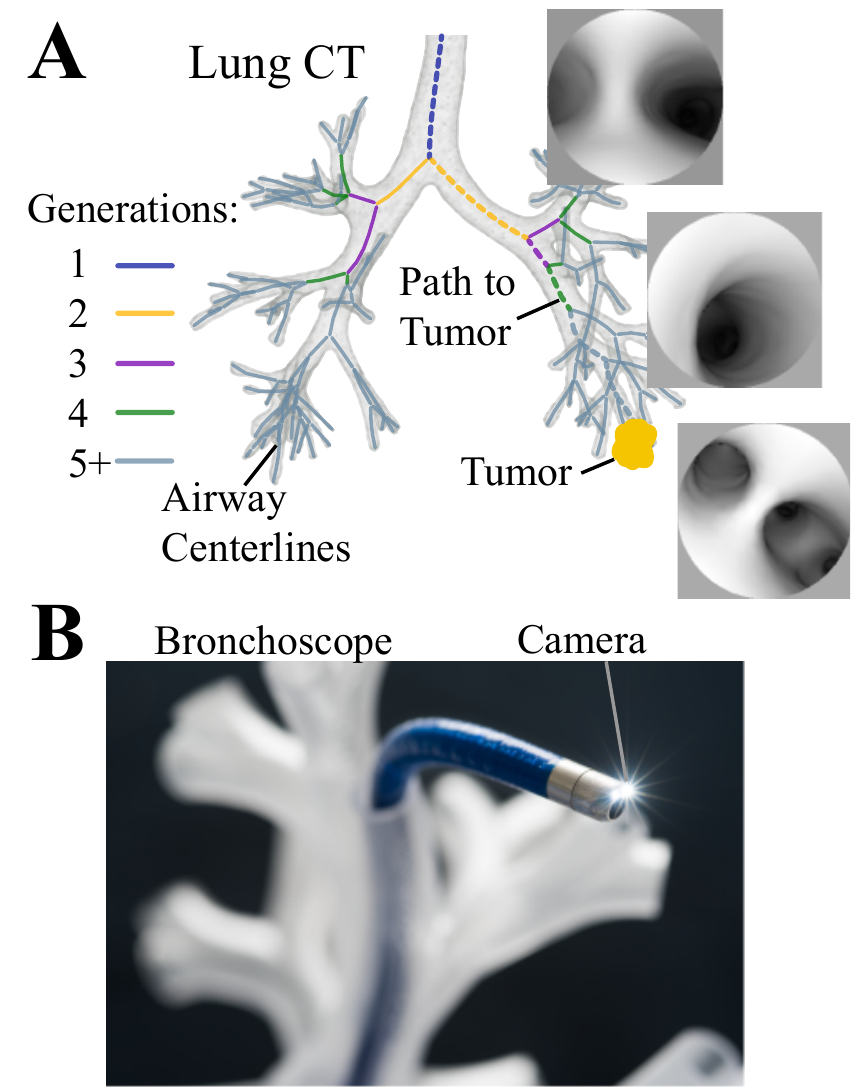}
  \caption{In A, a path through a preoperative lung CT is shown toward the site of a potential tumor. Images shown were rendered along the path to demonstrate what the physician would see as they move through the generations of the lung. The branching structure of the lung is represented in a lower dimensional skeletal tree based on the airway centerlines and the junctions between them, usually bifurcations. In B, a robotic bronchoscope (Auris Health, Inc.) is shown and the distal monocular camera used for visualizing the airways is highlighted.}
  \label{lung_task}
  \vspace*{-1mm}
\end{figure}

%
%
%
%
\IEEEPARstart{D}{iagnosing} lung cancer, the leading cause of cancer-related death world-wide, at an early stage significantly improves patient outcomes \cite{cruz2011lung}. Physicians biopsy potentially cancerous nodules in the lung by manually driving long, flexible bronchoscopes through the patient's airways, shown in Fig. \ref{lung_task}. This minimally invasive approach is preferred when the nodule is accessible, given the lower complication rates (2.2\% vs 20.5\%) compared to transthoracic needle biopsy \cite{ost2016diagnostic,dibardino2015transthoracic}.

Before the bronchoscopy, the physician selects biopsy targets in the lung's computed tomography (CT) scan. During the bronchoscopy, the physician maps the feedback from the bronchoscope (2D image) to the CT (3D map). This process is called localization \cite{wu2018image}. With accurate localization information, a physician can select the correct airways leading to the biopsy targets. 

Recently, robotic endoscopy systems have been developed to further aid the physician in reaching the target \cite{Rafii-Tari}. If the localization were sufficiently precise, a closed-loop control system could drive the bronchoscope without human intervention. Autonomous driving may improve localization by keeping clear view of the airways and bifurcations. It could also allow for de-skilling standard bronchoscopies, potentially reducing the cost of the procedure with a single pulmonologist monitoring multiple simultaneous procedures.

\begin{figure*}[thpb]
      \centering
      \includegraphics[width=0.7\textwidth]{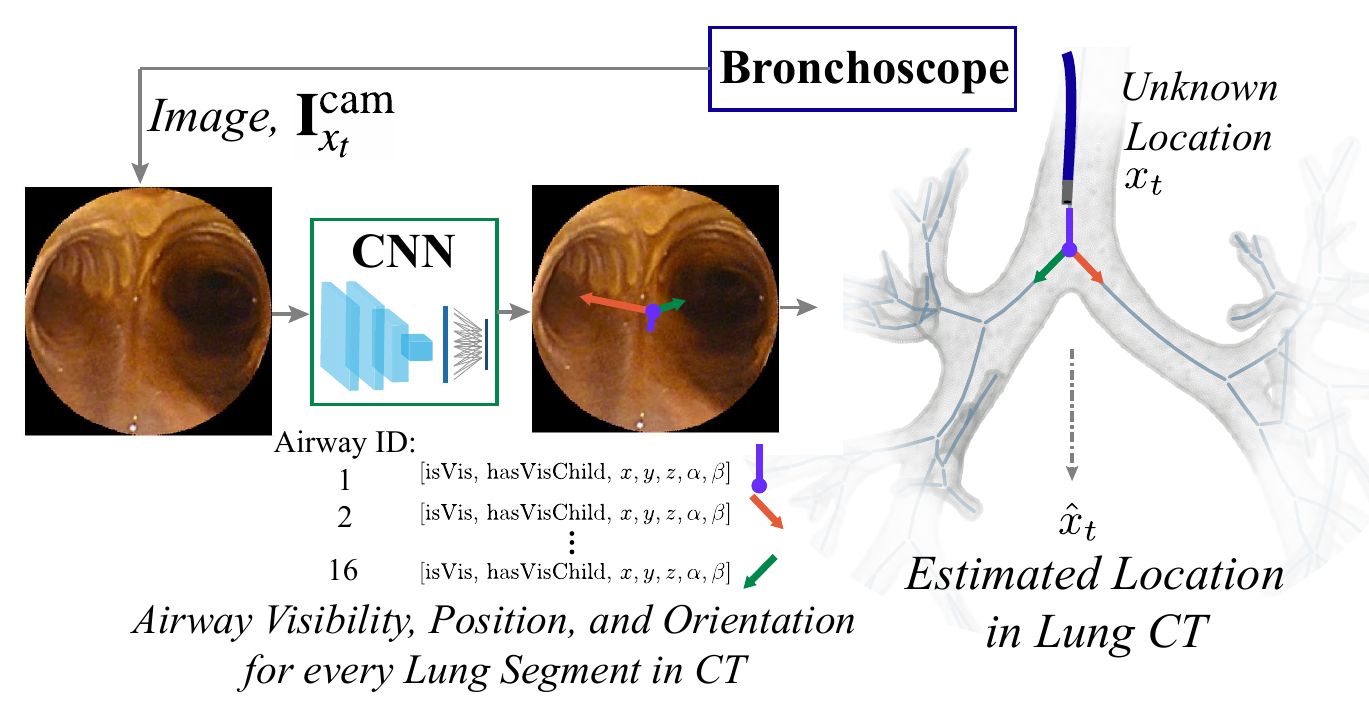}
      \caption{The inputs and outputs of AirwayNet are shown. A camera image from the bronchoscope's true position, $\mathbf{I}_{x_t}^{\text{cam}}$, passes through a trained CNN (ResNet-18) and outputs the airway characteristics for every airway in the lung skeleton. If the identified airway $hasVisChild$ and the children airways $isVis$ are consistent with the lung skeleton, the estimated position $\hat{x}_t$ is backed out from the measurement.}
      \label{cnn_architecture}
\end{figure*} 

Many sensing modalities can assist in localization, including a distal electromagnetic (EM) position sensor. After registration to the preoperative CT of the patient's chest, the position sensor can enable GPS-like directions on the road map to the target site \cite{rosell2012motion}. In these navigated bronchoscopies, the diagnostic yield varies across institutions, ranging from 67-74\% \cite{ost2016diagnostic,reynisson2014navigated}. Using electromagnetic navigation, the authors demonstrated closed-loop control of a robotic bronchoscope in a precisely registered phantom lung \cite{sganga2017orientation}. Additionally, techniques like fluoroscopy, radial ultrasound probes, and alternative endoscopic sensors (Raman spectroscopy, confocal, etc.) have been developed to further improve diagnostic yield \cite{reynisson2014navigated}. We chose to focus on advancing image-based approaches since cameras are the cheapest and most prevalent sensor for bronchoscopies, the local camera frame information is robust to breathing disturbances, and the information can be integrated with other modalities.

A localization algorithm must satisfy two requirements to aid decision-making: 1. it must accurately localize the bronchoscope and 2. it must operate in sufficiently real-time to enable closed-loop control. Several groups have developed image-based localization algorithms by comparing the bronchoscopic images to simulated images; however, these methods usually register images inefficiently at around 1-2 Hz and rely on highly realistic rendering \cite{Rai2008,Merritt2013}. Methods like SIFT and ORBSLAM have been used, but the airways have insufficient features and tracked features often drop out \cite{byrnes2014construction,visentini2017deep}. Anatomical landmarks have been tracked, like bifurcations \cite{shen2017branch}, lumen centers \cite{sanchez2017towards}, centerline paths \cite{Hofstad2014}, or similar image regions \cite{Luo2014}, but these approaches may not operate in real-time and tend make assumptions about the airway geometries using traditional computer vision techniques. 

Because of the difficulties traditional computer vision techniques face in this task, we decided to explore a deep learning approach. Using convolutional neural networks (CNN) to estimate the position and orientation of objects has been shown in many contexts, including for human posture and objects in a robotic hand \cite{Zhou_2016_CVPR,OpenAI2018}. The KITTI dataset is a large, high quality dataset to improve visual-based localization methods for autonomous cars \cite{geiger2012we}, and the top performing algorithms use variations of CNNs to process the visual information. In the lung domain, Visentini-Scarzanella \textit{et al.} and  used a CNN to estimate the depth map of 2D images in a phantom lung, which could then be registered to the 3D map, but localization is not reported \cite{visentini2017deep}. In our previous work, we used a CNN to localize a bronchoscope in real-time by predicting the offset between the camera image and a rendering at the expected position \cite{sganga2019offset}. This approach, called OffsetNet, showed 1.4 mm accuracy on a phantom lung sequence, but it fails to track other sequences.

In this work, we contribute two image-based deep-learning approaches, called  AirwayNet and BifurcationNet, that localize a bronchoscope in the lung CT frame. We evaluate the approaches on a dataset recorded from a silicone phantom lung. AirwayNet consistently tracked the bronchoscopic video in real time after training on simulated images generated from the preoperative scan of the phantom lung. Additionally, we demonstrate autonomous driving in the phantom lung using AirwayNet. After training on simulated images alone, the system reached 4 targets in the lung with a 95\% success rate. To the authors' knowledge, this is the first demonstration of image-based closed-loop control of a robotic bronchoscope. Finally, we evaluate AirwayNet on a dataset recorded from bronchoscopies in eight human cadaver lungs.
 
\vspace*{2mm}
\begin{table}[h]
\caption{Notation}
\label{notation_table}
\centering
\begin{tabular}{>{\centering\arraybackslash}p{1.5cm} p{6.25cm}}
\hline
\multicolumn{2}{c}{\textbf{Image Styles}}
 \\
 \rowcolor[HTML]{EFEFEF}
 $\mathbf{I}^{\text{cam}}$ & Image taken by a bronchoscope within the lung 
 \\
  $\mathbf{I}^{\text{sim}}$ & Image rendered by OpenGL using the lung CT 
 \\ 
 \rowcolor[HTML]{EFEFEF}
 $\mathbf{I}^{\text{rsim}}$ & $\mathbf{I}^{\text{sim}}$ image with varied rendering parameters and varied noise, smoothing and occlusions added \cite{Tobin2017}
 \\              
\hline
\multicolumn{2}{c}{\textbf{Error between True and Estimated Locations}}
\\
 \rowcolor[HTML]{EFEFEF}
$e_p$ & Position error (mm), defined as $e_p$ in \cite{Merritt2013}
\\   
$e_d$ & Direction angle error between pointing vectors, $p_z$, of the two views (\degree), defined as $e_d$ in \cite{Merritt2013}
\\
\rowcolor[HTML]{EFEFEF}
$e_r$ & Roll angle error between the $p_x$ axis after the $e_d$ was corrected for between views (\degree), defined as $e_r$ in \cite{Merritt2013}
\\              
\hline
\multicolumn{2}{c}{\textbf{CNN Outputs}}
\\
 \rowcolor[HTML]{EFEFEF}
$\mathbf{y}_{p}^{(i)}$ & Position vector including $x,y,z$ (mm), defined in camera frame of the furthest visible point on an airway, $i$
\\   
$\mathbf{y}_{d}^{(i)}$ & Direction vector including $\alpha, \beta$ (rad), defined in camera frame of the airway direction.
\\
\rowcolor[HTML]{EFEFEF}
$\mathbf{y}_{isVis}^{(i)}$ & True if any point along airway $i$'s centerline is visible\vspace*{0.5mm}   
\\         
$\mathbf{y}_{hasVisChild}^{(i)}$ & True if airway $i$'s bifurcation is visible  \vspace*{0.5mm}   
\\
\hline
\end{tabular}
\end{table}


\vspace*{-3mm}
\section{Method}
Shown in Fig. \ref{cnn_architecture}, at every step in the localization task, an image from the bronchoscope, $\mathbf{I}_{x_t}^{\text{cam}}$, at time $t$ from the position, $x_t$, is input to the localization algorithm. The algorithm outputs the set of visible airways and their positions and orientations, $\mathbf{y}_t$, with respect to camera frame. If a bifurcation is visible, airway information is used to calculate the 6 degree of freedom (DOF) location of the bronchoscope in CT frame, $\hat{x}_t$. AirwayNet is described in detail below, while BifurcationNet is described in the Appendix.


\subsection{Network Architecture}
AirwayNet consists of a deep residual convolutional network (CNN) and a single fully connected layer, which produces the output of size $500\times7$. The residual parts of our network implement the 18-layer architecture described in He \textit{et al.} \cite{he2016deep}. The CNN is implemented in Tensorflow, version 1.9 \cite{abadi2016tensorflow}. 

Each row of the CNN output corresponds to a unique airway, $i$, in the CT, denoted by “Airway ID” in Fig. \ref{cnn_architecture}, and the seven associated properties, $\mathbf{y}^{(i)}$. There are 2 visibility booleans, $isVis$ and $hasVisChild$. The measure $\mathbf{y}_{isVis}^{(i)}$ is true if any point along the airway centerline is visible, meaning the point lies within the field of view of the camera ($60^\circ$) and within the max visibility distance (set to 3 cm). The measure $\mathbf{y}_{hasVisChild}^{(i)}$ is true if the airway's bifurcation is visible. The remaining 5 properties describe the position and orientation of the airway in camera frame. AirwayNet regresses the position, $\mathbf{y}_p^{(i)}=(x,y,z)$, of the furthest point on the airway and its direction, $\mathbf{y}_d^{(i)}=(\alpha,\beta)$, in camera frame. The total number of rows is a hyperparameter set as an upper limit to the number of possible airways in the lung, which was set to 500 to keep the dimensions the same for each lung tested. 

For all experiments, AirwayNet models are trained on an NVIDIA Titan X GPU for 60k steps using Adam optimization ($\beta_1=0.9$, $\beta_2=0.999$, $\epsilon=10^{-8}$) to minimize a weighted L2 loss function on the airway positions and angles and a sigmoid cross entropy loss on the classification of the airways. 
\begin{align*}
  \mathcal{L} = \sum_{i=1}^M f (\mathbf{y}_{p}^{(i)}) \bigg(- & c_1 \cdot \mathbf{y}_{isVis}^{(i)} \log(\hat{\mathbf{y}}_{isVis}^{(i)})   \\
   - & c_2 \cdot \mathbf{y}_{hasVisChild}^{(i)} \log(\hat{\mathbf{y}}_{hasVisChild}^{(i)})   \\
   + & c_3 \cdot \mathbf{y}_{isVis}^{(i)}||\hat{\mathbf{y}}_{p}^{(i)} - \mathbf{y}_{p}^{(i)}||_2^2 \\
   + & c_4 \cdot \mathbf{y}_{isVis}^{(i)}||\hat{\mathbf{y}}_{d}^{(i)} - \mathbf{y}_{d}^{(i)}||_2^2 \bigg)
  \\
  f (\mathbf{y}_{p}^{(i)}) &= \max(c_5, c_6 - c_7 ||\mathbf{y}_{p}^{(i)}||_2 )
\end{align*}

The hyperparameters, $c$, are manually set to balance the classification and regression losses as well as to include a depth scaling to weigh the loss on nearby airways more heavily than distant airways. The regression losses are only incurred when airway $i$ is visible. The parameters set for all training in this paper are $c_1 = 2, c_2 = 2, c_3 = 1, c_4 = 10, c_5 = 0.1, c_6 = 6, c_7= 0.2$. The relationship between position and rotation errors are set according to a 1 mm:5.7\degree (10 mm:1 rad) \ ratio, which roughly relates to the fact that a 5.7\degree \ $e_d$ angle error results in an error of 1 mm for a location 10 mm in front of the camera.

As a comparison to AirwayNet, two other deep learning algorithms, OffsetNet and BifurcationNet, are evaluated on the phantom lung tracking test, shown in Fig. \ref{fig_tracking}. OffsetNet predicts the 6DOF offset between the bronchscopic image and the simulated image at the expected location in the lung \cite{sganga2019offset}. It is comprised of two 34-layer ResNets followed by a fully connected layer. BifurcationNet, described in detail in the Appendix, shares the same architecture of AirwayNet, but it outputs information about the visible airways without classifying them based on the CT. It relies on a novel particle filter to match the visible airways to airways in the CT. 

\subsection{Datasets}
\begin{figure}[thpb]
      \centering
      \hspace*{-1mm}
      \includegraphics[width=0.49\textwidth]{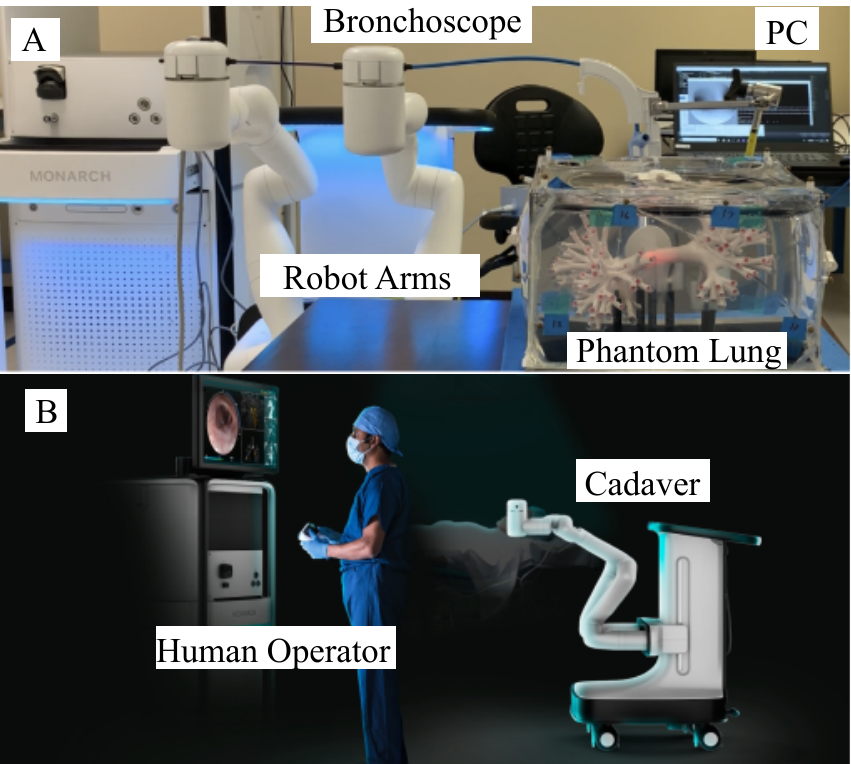}
      \caption{In A, the experimental setup for the phantom lung tests is shown. A PC laptop with with a 2.70 GHz CPU and 16 GB RAM ran the control loop in the driving experiments, including inference on the trained neural networks without GPU support. In B, a representative setup is shown for the cadaver experiments. Images courtesy Auris Health, Inc.}
      \label{fig_setup}
\end{figure}

\begin{figure}[thpb]
      \centering
      \hspace*{-1mm}
      \includegraphics[width=0.49\textwidth]{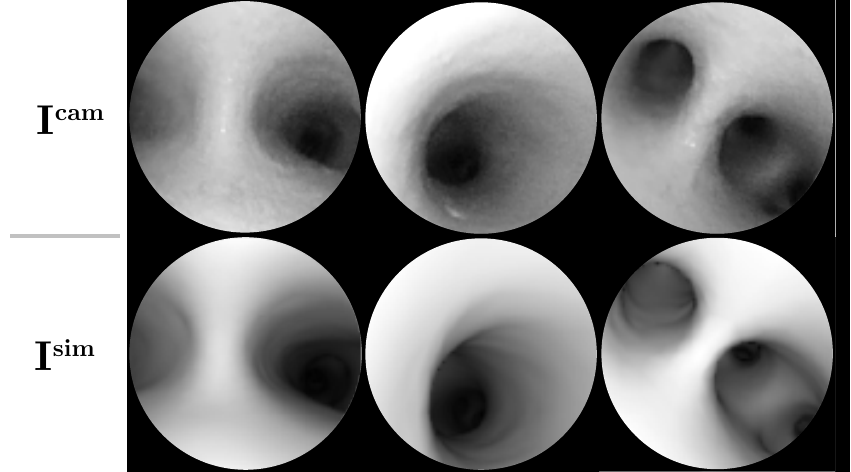}
      \caption{In the top row, example $\mathbf{I}^{\text{cam}}$ images taken within the phantom lung by the bronchoscope are shown. In the bottom row, the $\mathbf{I}^{\text{sim}}$ images are rendered at the same locations as the example $\mathbf{I}^{\text{cam}}$ images. Each image is grayscaled and normalized to zero mean and unit standard deviation before training and evaluation.}
      \label{fig_pics_koken}
\end{figure}

The $\mathbf{I}^{\text{cam}}$ images in the tracking experiment (Fig. \ref{fig_tracking}) are recorded by teleoperating a robotic bronchoscope (Monarch Platform, Auris Health Inc.) for 30 minutes in a phantom lung (Koken Co.), covering 3-8 generations of both lungs. The phantom lung is encased in silicone to better represent the material properties of the lung, shown in Fig. \ref{fig_setup}A. The images are manually registered to CT frame using the same method as \cite{sganga2019offset}. Example images are shown in Fig. \ref{fig_pics_koken}.

\begin{figure}[thpb]
      \centering
      \hspace*{-1mm}
      \includegraphics[width=0.49\textwidth]{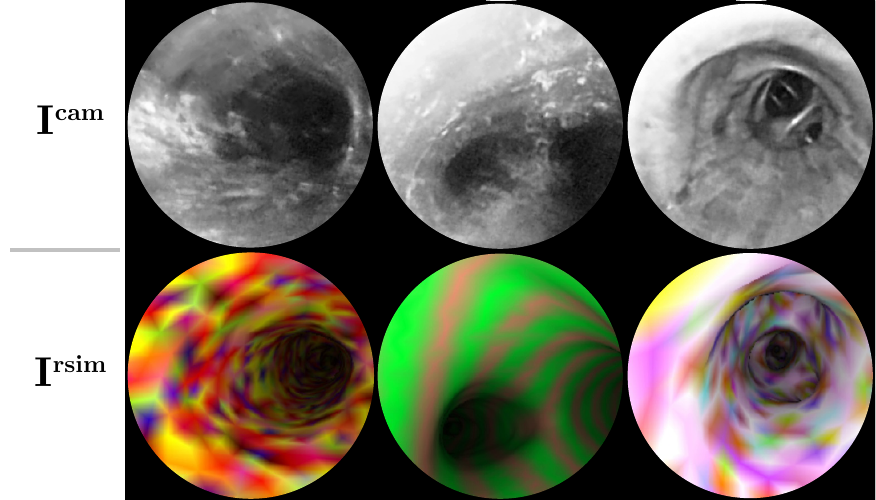}
      \caption{In the top row, example $\mathbf{I}^{\text{cam}}$ images taken within one of the eight human cadaver lung by the bronchoscope are shown. In the bottom row, the $\mathbf{I}^{\text{rsim}}$ images are rendered at the same locations as the example $\mathbf{I}^{\text{cam}}$ images. The $\mathbf{I}^{\text{rsim}}$ images are shown with patch and stripe randomization. These images are shown in color to highlight the effect, however they would be grayscaled and normalized before being used in training.}
      \label{fig_pics_cadaver}
\end{figure}

In the driving experiments (Fig. \ref{fig_driving}), the images are fed to AirwayNet in real-time. 

In the human cadaver localization experiments, the $\mathbf{I}^{\text{cam}}$ images are recorded by expert operators teleoperating a robotic bronchoscope (Monarch Platform, Auris Health Inc.) to reach predefined targets, shown in Fig. \ref{fig_setup}B. Up to two targets per side of the lung are registered for the localization experiments shown in Fig. \ref{fig_cadaver}. We exclude cadaver sequences in which the preoperative CT scan geometry deviated significantly from the geometry visible in the $\mathbf{I}^{\text{cam}}$ images. Example images are shown in Fig. \ref{fig_pics_cadaver}.

All $\mathbf{I}^{\text{rsim}}$ images used for training AirwayNet are rendered using PyOpenGL and a 3D lung STL from a segmented CT scan of the phantom lung (Monarch Platform, Auris Health Inc.) \cite{mansoor2015segmentation,sganga2019offset}. The rendering and domain randomization parameters match Sganga \textit{et al.} \cite{sganga2019offset}. Images are rendered at 60 Hz on a PC with no accelerations. All images were grayscaled and per-image normalized. 

For the tracking experiments in the phantom lung shown in Fig. \ref{fig_tracking}, each algorithm is trained on 202835 simulated images. The positions of the images are uniformly distributed in translation ($8$ mm diameter), direction ($\pm 23^\circ$), and roll ($\pm 23^\circ$) around the labeled positions of the recorded data. Five simulated images are generated around each labeled image position. 

In the driving experiment shown in Fig. \ref{fig_driving}, the same network trained for the tracking experiment is used. For the repeated driving test, shown in Fig. \ref{fig_repeated_driving}, a dataset of simulated images is generated for each target and a unique model is trained on each dataset. The positions of the images are distributed around the trajectory a simulated bronchoscope took as it navigated to each target. For targets 1-4, models are trained on 136250, 113500, 120000, and 122750 simulated images respectively. The image positions follow the same uniform distribution above.

When AirwayNet is trained on human cadaver lungs scans, 50 simulated images are generated around each labeled image position according to the same uniform distribution. Additional domain randomization is incorporated in the training sets to account for the lungs' textured appearance. The additional randomization consists of varying the surface color of the virtual lung in random patches and stripes. Fig. \ref{fig_pics_cadaver} shows examples of these randomizations before the images were grayscaled and normalized. For the patches, random RGB color variations are assigned to randomly sized and spaced sets of vertices in the underlying STL. For the stripes, a sine wave of randomized spatial frequency (uniform from 1 mm to 20 mm) radiating from a randomly selected lung vertex determines the magnitude of a random color to add to the lung vertices. Half of all simulated images are modified with color patches, and half are modified with stripes. Images can be modified with both color randomizations.

\subsection{Closed-loop Motion Control}
In the driving task, the planned trajectory to a given target consists of a list of airway ID's the robot is expected to see sequentially. The robot follows a given airway until it reaches within 1.5 cm of the airway's distal bifurcation and the next airway in the trajectory is visible. The motion controller consists of a proportional gain controller on the desired view angle ($\alpha$, $\beta$), which is calculated based on the projection of the target airway to a point 1.5 cm in front of the camera. This distance is a manually set hyperparameter. The insertion rate scales linearly based on the error in view angle. Once Target 1 is reached, the planned trajectory is adjusted for Target 2. This trajectory, however, starts at the trachea, like the first target's trajectory. To handle this situation and situations where the algorithm loses sight of the airways, the control loop reverses the insertion and aims the view angle towards the nearest visible airway until an airway on the new trajectory becomes visible.

\begin{figure}[thpb]
      \centering
      \hspace*{-6mm}
      \includegraphics[width=.5\textwidth]{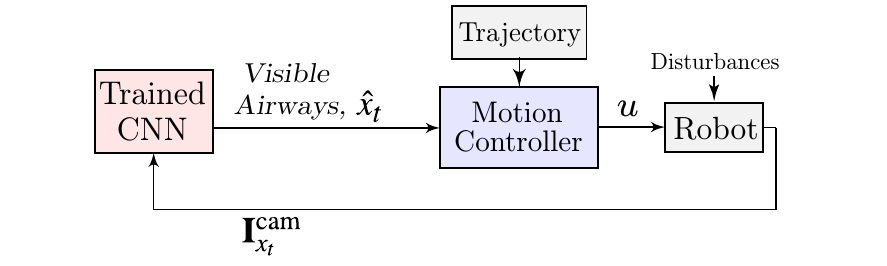}
      \caption{The driving control loop is shown, where a trained AirwayNet feeds the visible airways in a given image, $\mathbf{I}_{x_t}^{\text{cam}}$, into the motion controller. The motion controller compares the visible airways to the list of trajectory airways, and follows the nearest target airway using a proportional gain controller on the robot heading.}
      \label{control_loop}
\end{figure}

The motion controller is shown below. The proportional gain $k$ is set to 0.5, the max insertion velocity, $v_{ins}$, is set to 10 mm/s, and the max direction error, $e_{max}$, is set to $90^\circ$.

\begin{align*}
\left[\begin{array}{c} d\alpha\\d\beta \end{array}\right] &= \frac{1}{k}
R_{\theta} J du_{tendons}
\\
\left[\begin{array}{c} d\alpha\\d\beta \end{array}\right] &= \frac{1}{k} R_{\theta} \left[\begin{array}{cccc} 1 & 0 & -1 & 0 \\ 0 & 1 & 0 & -1 \end{array}\right] du_{tendons}
\\
du_{tendons} &= k J^{\dagger} R_{\theta}^\top \left[\begin{array}{c} d\alpha\\d\beta \end{array}\right] 
\\
du_{ins} &= f_{ramp}\left( \left\Vert \left[\begin{array}{c} d\alpha\\d\beta \end{array}\right] \right\Vert_2 \right)
\\
&= \max\left(0, v_{ins} \left(1 - \frac{1}{e_{max}}\left\Vert \left[\begin{array}{c} d\alpha\\d\beta \end{array}\right] \right\Vert_2 \right)\right)
\end{align*}

\section{Results}
On a laptop PC with a 2.70 GHz CPU, the localization algorithm ran at an average of 53.4 Hz, while the bronchoscope receives images at a rate of 25-30 Hz.

\subsection{Localization in Phantom Lung}
The tracking performance is measured along several dimensions to provide a detailed view of the how the localization would relate to driving decisions. Since successful navigation critically depends on identifying airways, the results emphasize the F1 score, which is the harmonic mean of precision and recall, on visible airways. The F1 score is broken down by each airway to show how performance is affected as the bronchoscope moves deeper into the branched structure. The precision-recall curves averaged over all airways provides a high-level sense of the algorithm's performance. The mean distance in position, direction, and roll between the labeled point, $x_t$, and the estimated position of the bronchoscope in CT frame, $\hat{x}_t$, is shown for when a bifurcation is visible and is correctly labeled. 

\begin{figure*}[thp]
      \centering
      \hspace*{-7mm}
      \includegraphics[width=1.\textwidth]{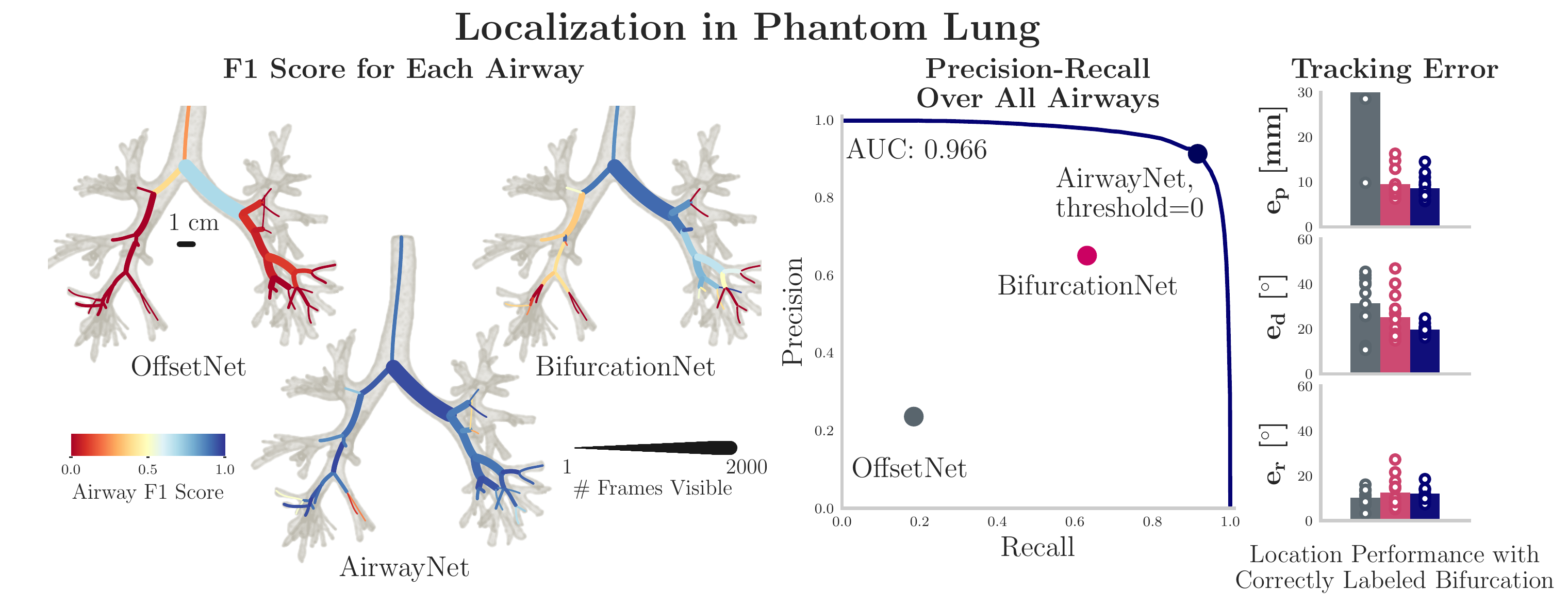}
      \caption{AirwayNet, BifurcationNet and OffsetNet, trained on $\mathbf{I}^{\text{rsim}}$, are shown in 13 independent tracking tasks in a phantom lung. Left, the F1 score in classifying airways is shown in color for each airway the bronchoscope saw. The number of frames with each airway visible is represented in the line thickness. Middle, the precision-recall curve is shown averaged across all airways. Right, the tracking analysis shows the error in position, direction and roll for the frames when the airways of a bifurcation were correctly labeled by the algorithm.}
      \label{fig_tracking}
\end{figure*}

In Fig. \ref{fig_tracking}, three deep learning algorithms are evaluated on 13 recorded bronchoscopic videos through the phantom lung. Each video starts in the trachea and ends at a target airway four to eight generations deep. As a comparison to AirwayNet, we show the performance of OffsetNet and BifurcationNet. Because OffsetNet and BifurcationNet do not assign relative probability to the airways, single precision and recall values are shown rather than the full precision-recall curve. AirwayNet evaluated with a threshold of 0 on its classification outputs is highlighted for comparison. On all metrics, including precision and recall on airways and the tracking errors $e_p$ and $e_d$, AirwayNet outperforms OffsetNet and BifurcationNet.

\subsection{Closed-Loop Control in Phantom Lung}
\begin{figure*}[thpb]
      \centering
      \hspace*{-12mm}
      \includegraphics[width=1.\textwidth]{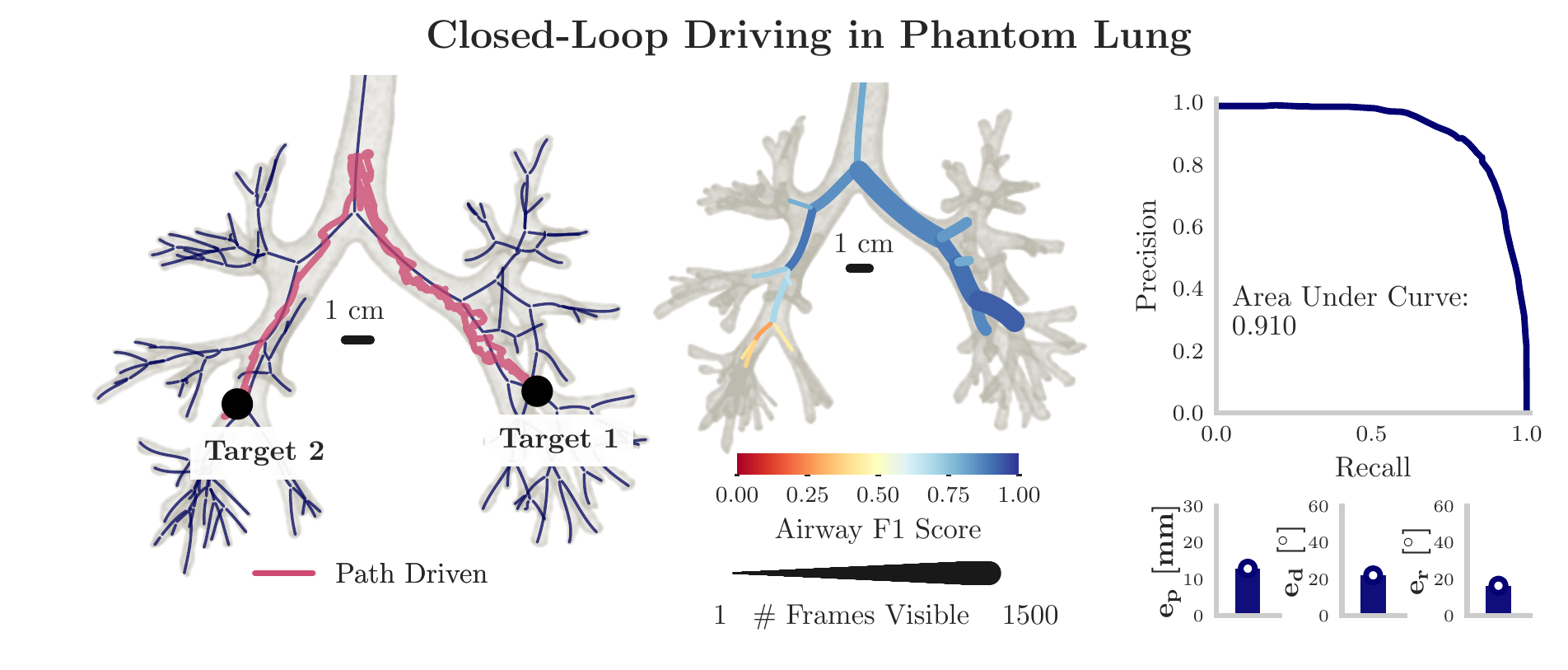}
      \caption{Closed-loop control of the robotic bronchoscope is shown using AirwayNet, trained on $\mathbf{I}^{\text{rsim}}$. The robot drives to two targets sequentially. Left, the position of the robot at each time step and the two targets it reached. The entire sequence lasted 79 seconds, and the control loop operated at 48 Hz. Center, the same analysis as Fig. \ref{fig_tracking} is shown.}
      \label{fig_driving}
\end{figure*}

A trained AirwayNet enables closed-loop control of the robot, shown in Fig. \ref{fig_driving}. For the driving task, two targets are selected for the robot to reach sequentially. After reaching the first target, the robot reverses until it sees the next set of target airways. The path driven to both targets is shown in Fig. \ref{fig_driving}, as well the same analysis performed in Fig. \ref{fig_tracking}. The control loop for this task is shown in Fig. \ref{control_loop}.  The model used for this experiment is the same as the AirwayNet model evaluated in Fig. \ref{fig_tracking}. The loop operates at 48 Hz for this experiment. 

To evaluate the consistency of the driving performance, four targets are selected to reach in repeated trials. The robot starts each trial in the trachea, and five trials are run sequentially on each target before moving to the next. The experiments are run on the same bronchoscope on the same day. The robot successfully reaches the target on 19 out of 20 trials. In three successful trials, the robot recovers after losing sight of the target airway. In these situations, the robot recovers by backing up until it recognizes airways on the trajectory again. The single failure case involves a failed recovery attempt. When attempting to reach target three, the robot runs too close to the wall, loses sight of the airway, and after backing up, runs back into the wall.

\begin{figure}[thpb]
      \centering
      \hspace*{-12mm}
      \includegraphics[width=.5\textwidth]{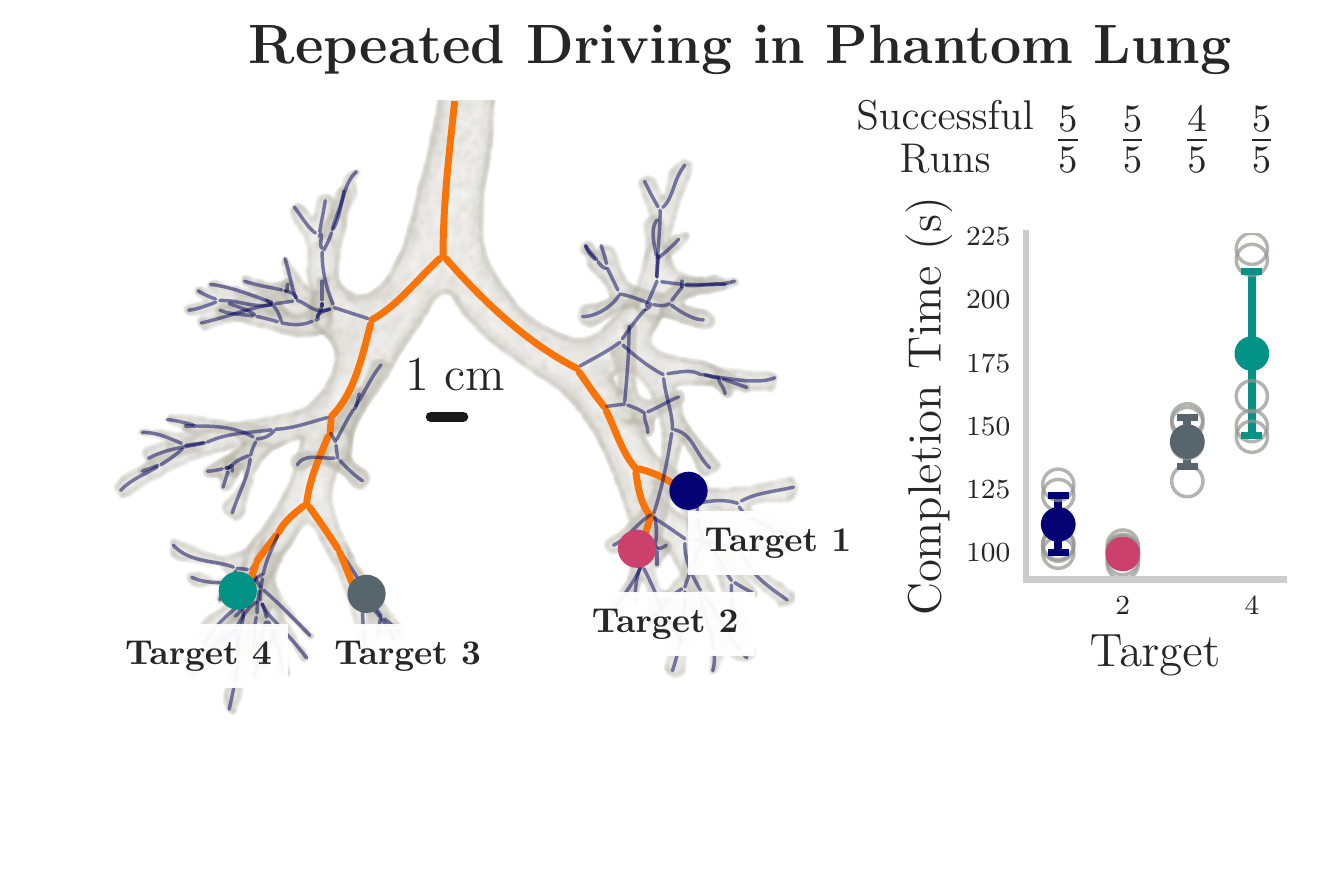}
      \vspace*{-12mm}
      \caption{Repeated closed-loop control tests of the robotic bronchoscope is shown using AirwayNet, trained on $\mathbf{I}^{\text{rsim}}$. A unique AirwayNet was trained for each target. The robot drives to each target 5 times sequentially, starting in the trachea each time. Left, the centerline trajectories and terminal points for each target are shown. Right shows the number of successful runs and the completion time for each successful run. The mean and standard deviation of completion time are highlighted. The control loop operated at 10 Hz.}
      \label{fig_repeated_driving}
\end{figure}

\subsection{Localization in Human Cadaver Lungs}
\begin{figure*}[thpb]
      \centering
      \hspace*{-12mm}
      \includegraphics[width=1.\textwidth]{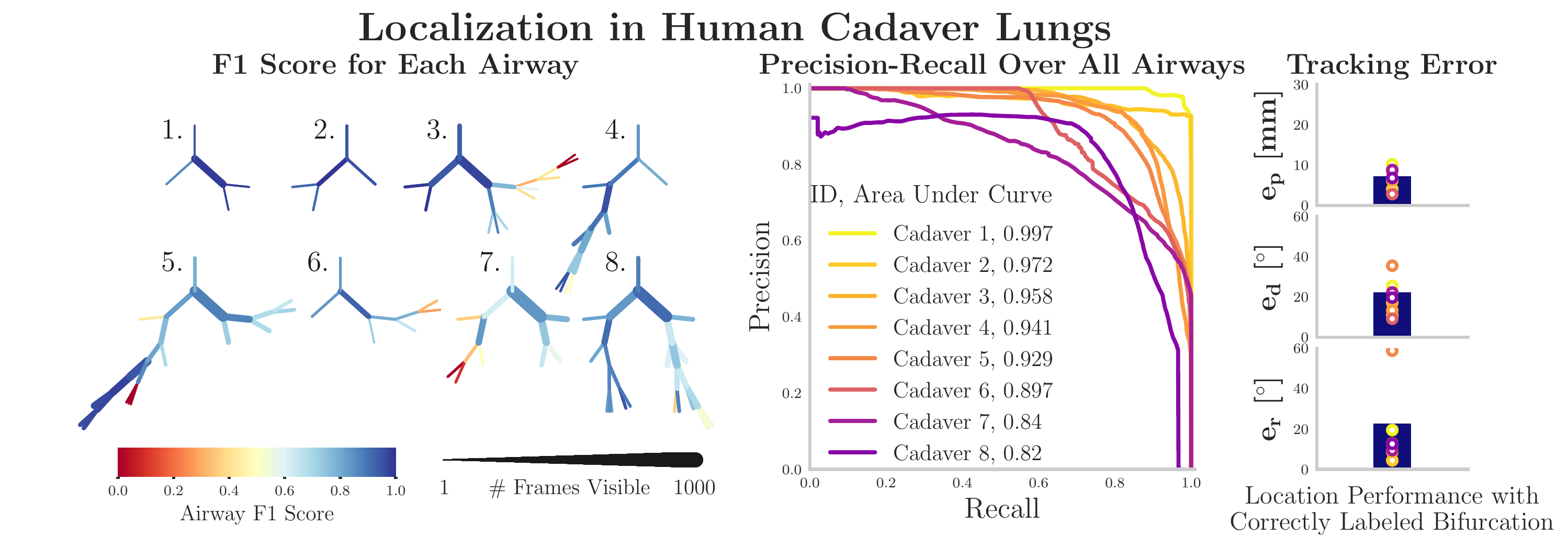}
      \caption{AirwayNet, trained on $\mathbf{I}^{\text{rsim}}$, is shown in 13 independent tracking tasks in eight cadaver lungs. Left, the F1 score in classifying airways is shown in color for each airway the bronchoscope saw. The number of frames with each airway visible is represented in the line thickness. Right, the precision-recall curve is shown averaged across all airways. Below, the tracking analysis shows the error in position, direction and roll for the frames when the airways of a bifurcation were correctly labeled by the algorithm.}
      \label{fig_cadaver}
\end{figure*}

Similar to the localization task in Fig. \ref{fig_tracking}, AirwayNet's tracking performance is evaluated on 19 recorded sequences in eight human cadaver lungs, shown in Fig. \ref{fig_cadaver}. The areas under the curve for each of the cadavers range from 0.82 to 0.997. The performance varies between cadavers and targets. Despite the difference in domain from the simulated images, $\mathbf{I}^{\text{sim}}$, AirwayNet successfully identifies the majority of airways in the lungs. In Table \ref{randomization_table}, the area under the curve for each cadaver is shown for models trained with and without domain randomization. Averaged across the eight lungs, the domain randomization increased the areas under the curved by 0.17.

\vspace*{2mm}
\begin{table}[h]
\caption{Areas under the precision-recall curves for each cadaver are shown comparing AirwayNet trained on images with domain randomization, $\mathbf{I}^{\text{rsim}}$, to AirwayNet trained on images without randomization, $\mathbf{I}^{\text{sim}}$.}
\label{randomization_table}
\centering
\begin{tabular}{>{\centering\arraybackslash}p{0.75cm} p{0.5cm} p{0.5cm} p{0.5cm} p{0.5cm} p{0.5cm} p{0.5cm} p{0.5cm} p{0.5cm}}
\hline
& \multicolumn{8}{c}{\textbf{Areas under the Precision-Recall Curves for each Cadaver}}
 \\
 \rowcolor[HTML]{EFEFEF}
 & 1 & 2 & 3 & 4 & 5 & 6 & 7 & 8
 \\
 $\mathbf{I}^{\text{sim}}$ & 0.985 & 0.836 & 0.883 & 0.548 & 0.779 & 0.676 & 0.574 & 0.682 \\
 \rowcolor[HTML]{EFEFEF}
 $\mathbf{I}^{\text{rsim}}$ & 0.997 & 0.972 & 0.958 & 0.941 & 0.929 & 0.897 & 0.840 & 0.820 \\
\hline
\end{tabular}
\end{table}

\vspace*{-1mm}
\section{Discussion}
This work demonstrates that training a deep neural network on simulated images alone can enable autonomous driving of a bronchoscope in a phantom lung. AirwayNet tracks bronchoscopic videos in a phantom lung and in human cadaver lungs accurately and in real-time. Given the similarity of airways, the result that a simple classifier with no contextual information can consistently identify airways is unexpected. The performance of this algorithm is sufficiently precise and real-time to allow for autonomous driving. It also demonstrates how domain randomization can bridge the domain gap between simulated images and human cadaver lung images. 

As shown in Fig. \ref{fig_tracking}, the performance of three deep learning architectures varies widely. AirwayNet outperforms both OffsetNet and BifurcationNet. OffsetNet's fragility was discussed by the authors in \cite{sganga2019offset}. It relies heavily on its expected position. When that expectation deviates from the true position, its estimates fail to recover on average. BifucationNet combines airway feature extraction and knowledge of its state in a particle filter to better handle the task. BifurcationNet's formulation matches the authors' intuitive understanding of how humans navigate the lung, and while it outperforms OffsetNet, it performs worse than AirwayNet. BifurcationNet struggles due to two main reasons. First, BifurcationNet's particle filter is sensitive to estimates from previous frames, so incorrectly classifying airways makes future predictions more difficult. AirwayNet, on the other hand, does not require a particle filter and does not factor previous estimates into the current estimate, so mistakes are limited to the specific frame. Second, to classify an airway, AirwayNet can leverage all the information in image space, while BifurcationNet uses only 7 airway characteristics plus insertion and state history. The image space information evidently makes up for the seemingly harder task of classifying airways without prior context.

As shown in Fig. \ref{fig_tracking}, AirwayNet begins to struggle in later generations. We believe two factors contribute to this: the CT scan's fidelity decreases deeper in the lung because the termination of airways creates visual artifacts, and we trained the network on fewer images in the periphery. One approach that can improve the network's performance in the periphery is to limit its training set to a sparse set of peripheral airways. In lung cancer biopsies, the target is known preoperatively; therefore, a model can be trained to a specific target, which decreases the complexity of the problem. This type of training was implemented in the repeated driving experiment, shown in Fig. \ref{fig_repeated_driving}. 

Given the tracking performance in the phantom lung, we chose to pursue autonomous driving with AirwayNet. Using the same model shown in Fig. \ref{fig_tracking}, the system is able to identify the airways along its planned trajectory and command the heading angle toward the airway centerlines. To demonstrate the flexibility of the system, a second target is selected when the robot successfully reaches the first target. The second target requires the robot to retract through the lung until it identifies an airway on its new trajectory. While Fig. \ref{fig_driving} shows the robot successfully reach two distant targets, the analysis of the localization performance shows slightly worse localization performance than in Fig. \ref{fig_tracking}. This may be related to label fidelity and the robot driving to poses that were less frequently simulated in training. 

To evaluate driving consistency, we ran a repeated driving experiment, shown in Fig. \ref{fig_repeated_driving}. By reaching the target in 19 of 20 trials, the system demonstrates it can reliably reach targets in the phantom lung. The network architecture's ability to recover from lost localization is demonstrated in the three trials that involve the robot backing out of a particular orientation with no localization information. Given the stochasticity of the network output, each trial involves a slightly different paths, which is why the recovered robots can successfully reach targets on their second attempts. The control loop unexpectedly ran at only 10 Hz on these trials. The firmware on the robot was different than the firmware used for the experiment in Fig. \ref{fig_driving}, and this likely affected the speed, as the localization and motion control algorithms were shown to run at $>$50 Hz in other experiments.

Extending the results from the phantom lung to human cadaver lungs requires updating the domain randomization to include coloring the lung with random patches and stripes. The results in cadavers provides evidence that this technique may be used for driving in cadavers and potentially live humans. In fact, live human lungs are often healthier and their appearance may match the simulated images better, which would improve the network's performance. 

There are several limitations to this approach. AirwayNet is highly dependent on the the CT scan's fidelity to the underlying anatomy. While it demonstrates the ability to handle the deformation caused by controlled respiration in the cadavers, significant deformations would likely impact the algorithm's accuracy. The algorithm would also struggle when the view is occluded due to fluid or collapsed peripheral airways. Additionally, the variance between cadavers is a concern when deploying the algorithm for autonomous driving in cadavers and live pigs. The algorithm demonstrates the ability to recover from situations with limited vision, but the localization can be made more robust by integrating information from other sensors like the insertion depth and EM position sensors.

This work demonstrates the feasibility of autonomous control of a minimally invasive surgical robot. Autonomy may offer advantages in surgery beyond matching human performance by allowing the physician to focus on higher level decision making and enabling physicians to monitor operations on multiple patients in parallel \cite{fagogenis2019autonomous}.

\section{Conclusion}
This work introduces a novel deep learning localization algorithm, evaluates it in a phantom lung and human cadaver lungs, and demonstrates autonomous driving in the phantom lung. Future work will investigate the localization performance in live human data and driving performance in human cadavers and live pigs.


%

\appendix[BifurcationNet Overview]

\begin{figure*}[thpb]
      \centering
      \includegraphics[width=0.9\textwidth]{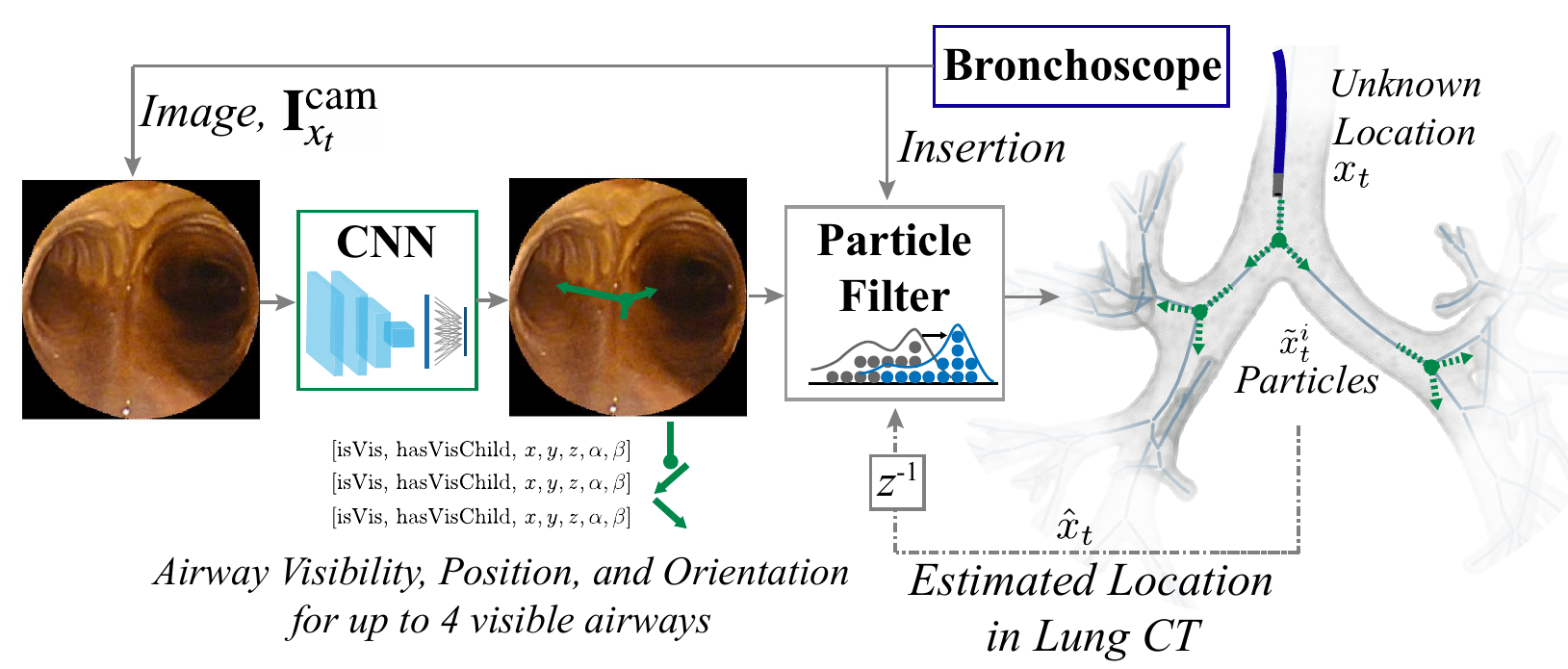}
      \caption{The control loop for BifurcationNet is shown. A camera image from the bronchoscope's true position, $I_(x_t)^{cam}$, at time t passes through a trained CNN (ResNet-18) and outputs a matrix, representing the characteristics of 4 airways. Two booleans capture the airway's visibility and the airway's bifurcation visibility. The CNN also outputs the camera frame position $(x,y,z)$ in mm of the furthest visible point of the airway, which corresponds to the bifurcation when it is visible, and $(\alpha,\beta)$ representing the $XYZ$ Euler angles for camera frame angle about X and Y, respectively. The 4 airways are ordered based on the position proximity to the camera and angle of the airway. The particle filter compares how this measurement relates to the most likely bifurcations based on the previous state and the current insertion from the robot, finding the posterior probability of each particle. The bifurcation with the highest posterior probability is selected and used to calculate $\hat{x}_t$. This is fed into the particle filter on the next time step.}
      \label{bifnet_architecture}
\end{figure*}

Shown in Fig. \ref{bifnet_architecture}, at every step in the localization task, an image from the bronchoscope, $\mathbf{I}_{x_t}^{\text{cam}}$, at time $t$ from the position, $x_t$, and the current absolute robot insertion (mm), $i_t$, is provided to the localization algorithm and the algorithm outputs a 6 degree of freedom (DOF) location estimate in CT frame, $\hat{x}_t$, along with the set of visible airways and their positions and orientations, $\hat{y}_t$, with respect to camera frame. A particle filter assigns the camera frame airways, $\hat{y}_t$, to the CT airways, $\hat{\mathbf{y}}_t$.

Similar to AirwayNet, BifurcationNet identifies a set of common characteristics for the visible airways in each image. Each airway is characterized through 2 visibility measures, $isVis$ if the airway is visible and $hasVisChild$ if the airway's bifurcation is visible. For each airway, the networks regress the camera frame position $(x,y,z)$ of the furthest point on the airway and its angle $(\alpha,\beta)$ . BifurcationNet outputs the airway characteristics of up to four visible airways into a novel particle filter that classifies the airways based on the most probable airways in the CT map, shown in Fig. \ref{bifnet_architecture}.

Bifurcation net shares the 18-layer Resnet architecture, like AirwayNet, and it is trained using the same Adam optimization and loss function. 

\subsection*{Particle Filter}
The key feature of BifurcationNet is the novel particle filter used to pair the visible airways, $y$, with the airways in the underlying CT, $\mathbf{y}$. When a bifurcation is visible and at least two children airways are visible, the particle filter determines the probability the measurement fits a given CT bifurcation, $\tilde{\mathbf{y}}^{(i)}$. The particle filter also calculates the prior probability of seeing that CT bifurcation and compares it to $n$ other candidate bifurcations. For all experiments shown, 3 particles were used. With the measurement probability and the prior probability, the filter calculates the posterior probability of the bronchoscope seeing a given particle bifurcation,  $p(x_t | \tilde{\mathbf{y}}^{(i)}, \hat{y}_t)$. In the work shown here, three bifurcations with the highest prior probability were compared. The filter assigns the bifurcation with the maximum posterior probability to the visible airways and uses that assignment to calculate the estimated bronchoscope position, $\hat{x}_t$. Each step is detailed below.

\textit{Measurement Probability}: the probability an observation matches a given bifurcation in the CT, $p_{fit}$, is based on how well the children airways align with the expected airways once the bifurcation point and the parent airway direction is aligned with the CT. To resolve the roll about the parent axis, the different child airway assignments to the underlying CT bifurcation were permuted and compared. For a given airway assignment, the optimal roll about the parent axis was calculated by minimizing the weighted average of the airway angle offsets. The probability of the fit is calculated based on a Gaussian distribution over the cosine angle difference between the measured child airway directions and CT airway directions. 

\begin{align*}
\text{Let } p(x; \mu,\sigma) &= \frac{1}{\sqrt{2\pi\sigma^2}}e^{-\frac{(x-\mu)^2}{2\sigma^2}}) 
\\
p_{fit} &= \frac{1}{n} \sum_i p(\tilde{\mathbf{y}}_d^{(i)\top} \hat{y}_d^{(i)}; 0, \sigma_{fit})
\end{align*}

\textit{Prior}: the bifurcation prior is calculated based on the current robot insertion length and the filter's previous state. The insertion probability, $p_{ins}$, is based on the distance between a given CT bifurcation, $i$, length from the trachea, $\tilde{z}^{(i)}_{bif}$, and the robot insertion, $u_{ins}$, plus the observed bifurcation depth in camera frame, $\hat{y}_{p_z}^{(i)}$. The prior airway probability, $p_a$, increases the probability of bifurcations near previously visible airways. If the bifurcation is 1, 2, or 3 generations removed from a visible airway, $d(\mathbf{y}^{(i)},\mathbf{y}^{(j)})$, its relative probability increments by 1, 0.1 and 0.01. For $p_x$, the previously estimated 3DOF location, $\hat{x}_{t-1}$, is used to calculate a 3D multivariate Gaussian with the particle estimate. Additionally, a roll probability is included to prevent sudden rotations and the previous roll value.

\begin{align*}
p_{prior} &= p_{ins} p_a p_x p_r
\\
p_{ins} (\tilde{z}^{(i)}_{bif} | u_{ins}, \hat{y}) &= p(u_{ins} + \hat{y}_{p_z}^{(j)} - \tilde{z}^{(i)}_{bif}; 0, \sigma_{ins})
\\
p_{a} (\tilde{\mathbf{y}}^{(i)} | \hat{\mathbf{y}}_{t-1}) &= \frac{\sum_j p_{gen}(\hat{\mathbf{y}}^{(j)}_{t-1}, \tilde{\mathbf{y}}^{(i)})}{\sum_i \sum_j p_{gen}(\hat{\mathbf{y}}^{(j)}_{t-1}, \tilde{\mathbf{y}}^{(i)})}
\\
p_{gen} (\mathbf{y}^{(i)}, \mathbf{y}^{(j)}) &=  \begin{cases}
10^{1 - d(\mathbf{y}^{(i)}, \mathbf{y}^{(j)})} \text{ if } d(\mathbf{y}^{(i)}, \mathbf{y}^{(j)}) \leq 3,\\
0 \text{ else}
\end{cases}
\\
p_x (\tilde{x}_p^{(i)} | \hat{x}_{p, t-1}) &= p(\|\tilde{x}_p^{(i)} - \hat{x}_{p, t-1}\|_2; 0, \sigma_x)
\\
p_r (\tilde{x}_{r}^{(i)} | \hat{x}_{r, t-1}) &= p(|\tilde{x}_r^{(i)} - \hat{x}_{r, t-1}|; 0, \sigma_r)
\end{align*}

The posterior probability for a given bifurcation, $i$, was then determined and the bifurcation with maximum likelihood would be selected as the estimate.
\begin{align*}
p(\tilde{\mathbf{y}}^{(i)} | \hat{y}_t, \hat{\mathbf{y}}_{t-1}, \hat{x}_{t-1}) &= p_{fit} p_{prior}
\\
\hat{x}_{t}, \hat{\mathbf{y}}_{t} &= \max_i p(\tilde{\mathbf{y}}^{(i)} | \hat{y}_t, \hat{\mathbf{y}}_{t-1}, \hat{x}_{t-1})
\end{align*}


\section*{Acknowledgment}
The authors would like thank Auris Health Inc. for the equipment and support, the NIH Biotechnology Training Grant and Bio-X for support, and NVIDIA for the Titan X GPU.

\ifCLASSOPTIONcaptionsoff
  \newpage
\fi

\begin{IEEEbiography}[{\includegraphics[width=1in,height=1.25in,clip,keepaspectratio]{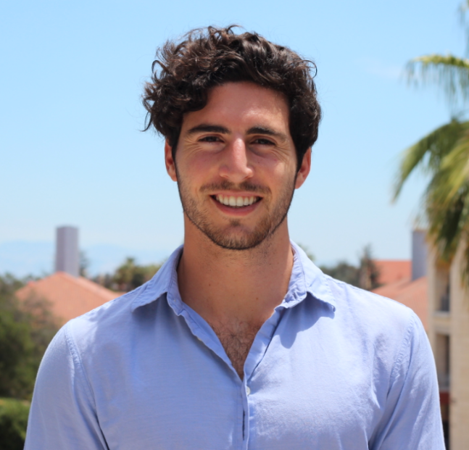}}]{Jake Sganga}
 earned a B.S. in Biomedical Engineering from Duke University and an M.S. in Bioengineering from Stanford University. His research focuses on the design of algorithms for the control and localization of soft surgical robotics.
\end{IEEEbiography}

\begin{IEEEbiography}[{\includegraphics[width=1in,height=1.25in,clip,keepaspectratio]{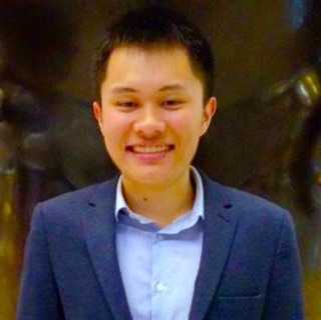}}]{David Eng}
 earned a B.S. and M.S. in Computer Science from Stanford University.
\end{IEEEbiography}


\begin{IEEEbiography}[{\includegraphics[width=1in,height=1.25in,clip,keepaspectratio]{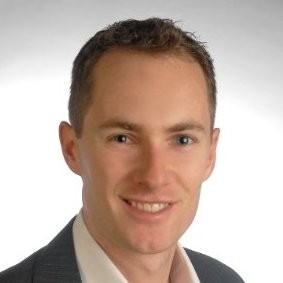}}]{Chauncey Graetzel}
 earned a B.Sc and M.Sc in Microtechnology from the Swiss Federal Institute of Technology in Lausanne (EPFL), and a Ph.D. in Bio-microrobotics from ETH Zurich. He was previously Head of Research at Optotune, bringing novel robotic optical devices to market. At Auris Health, he is Sr. Manager of Robotics \& Controls, where he has spearheaded the development and first commercial release of key robotic algorithms controlling the medical devices.
\end{IEEEbiography}

\begin{IEEEbiography}[{\includegraphics[width=1in,height=1.25in,clip,keepaspectratio]{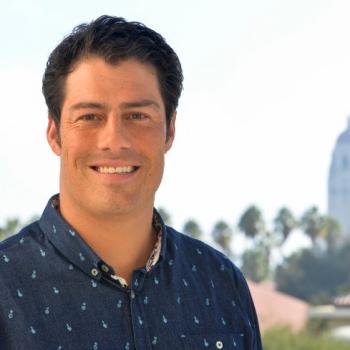}}]{David B. Camarillo}
 is Assistant Professor of Bioengineering, (by courtesy) Mechanical Engineering and Neurosurgery at Stanford University. Dr. Camarillo holds a B.S.E in Mechanical and Aerospace Engineering from Princeton University, a Ph.D. in Mechanical Engineering from Stanford University and completed postdoctoral fellowships in Biophysics at the UCSF and Biodesign Innovation at Stanford.
\end{IEEEbiography}




\end{document}